\documentclass[letterpaper]{article} 
\usepackage{aaai2026}  
\usepackage{times}  
\usepackage{helvet}  
\usepackage{courier}  
\usepackage[hyphens]{url}  
\usepackage{graphicx} 
\usepackage{natbib}  
\usepackage{caption} 
\usepackage{algorithm}
\usepackage{algorithmic}
\usepackage{nicefrac}       
\usepackage{microtype}      
\usepackage{xcolor}         
\usepackage{pifont}
\usepackage{booktabs}
\usepackage{pifont}
\usepackage{multirow}
\usepackage{amsfonts}       
\usepackage{threeparttable}
\usepackage{newfloat}
\usepackage{listings}
\DeclareCaptionStyle{ruled}{labelfont=normalfont,labelsep=colon,strut=off} 
\floatstyle{ruled}
\newfloat{listing}{tb}{lst}{}
\floatname{listing}{Listing}
\title{Towards Unified Vision Language Models for Forest Ecological Analysis in \\Earth Observation}
\author{
    Xizhe Xue\textsuperscript{\rm 1,}\textsuperscript{\rm 2}, Xiao Xiang Zhu\textsuperscript{\rm 1,}\textsuperscript{\rm 2}\thanks{Corresponding Author},
}
\affiliations{
    \textsuperscript{\rm 1}Technical University of Munich\\
    \textsuperscript{\rm 2}Munich Center for Machine Learning\\

    xizhe.xue@tum.de, xiaoxiang.zhu@tum.de
}

\usepackage{bibentry}

\begin{document}

\maketitle

 \begin{abstract}
Recent progress in vision language models (VLMs) has enabled remarkable perception and reasoning capabilities, yet their potential for scientific regression in Earth Observation (EO) remains largely unexplored. Existing EO datasets mainly emphasize semantic understanding tasks such as captioning or classification, lacking benchmarks that align multimodal perception with measurable biophysical variables. To fill this gap, we present REO-Instruct, the first unified benchmark designed for both descriptive and regression tasks in EO. REO-Instruct establishes a cognitively interpretable logic chain in forest ecological scenario: human activity → land-cover classification → ecological patch counting → above-ground biomass (AGB) regression, bridging qualitative understanding and quantitative prediction. The dataset integrates co-registered Sentinel-2 and ALOS-2 imagery with structured textual annotations generated and validated through a hybrid human–AI pipeline. Comprehensive evaluation protocols and baseline results across generic VLMs reveal that current models struggle with numeric reasoning, highlighting an essential challenge for scientific VLMs. REO-Instruct offers a standardized foundation for developing and assessing next-generation geospatial models capable of both description and scientific inference.

\end{abstract}

\begin{links}
    \link{Code}{https://github.com/zhu-xlab/REO-Instruct}
\end{links}

\section{Introduction}

Earth Observation (EO) data has become a essential resource for diverse scientific research, crucial in areas such as ecological monitoring~\cite{allenm3leo}, weather forecasting~\cite{nguyen2024climatelearn}, disaster response and population dynamics analysis~\cite{batista2020uncovering}. These researches require not only a thorough understanding of image content but also precise regression of the real-world scientific attributes they represent~\cite{benson2024multi,hamilton2024combining}. Regression, in this context, involves modeling the relationship between environmental metrics and relevant features, enabling accurate predictions and deeper insights into complex geospatial or  ecological processes.

\begin{figure}
    \centering
    \includegraphics[width=1.0\linewidth]{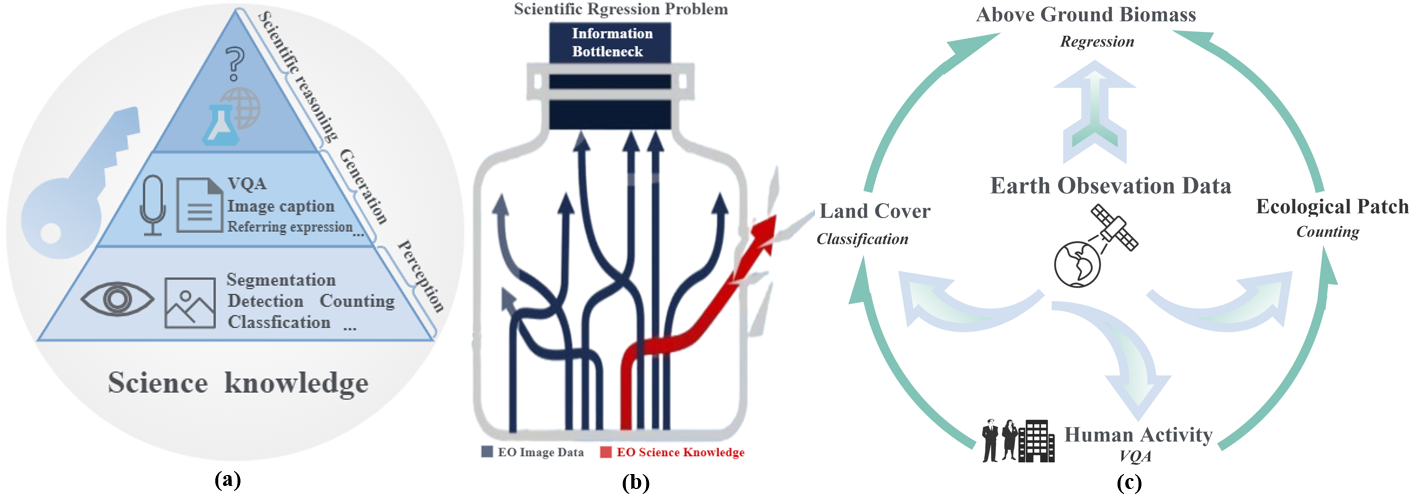}
\caption{ 
\textbf{(a). Hierarchical structure of VLM capabilities}: From basic perception tasks to higher-order reasoning tasks;
\textbf{(b). Advantages of VLM for EO regression tasks}: By integrating scientific domain knowledge with EO image data, VLMs overcome the information bottleneck of traditional image-only regression models, enabling deeper insights and improved scientific reasoning;
\textbf{(c). Interplay between regression and generation tasks}: Using AGB estimation as an example, the intrinsic link between regression and generation targets allows collaborative processing in a unified framework, enhancing prediction accuracy and reliability.}
    \label{fig:motivation_fig}
\end{figure}

\begin{table*}
\centering
\caption{Comparison of recent methods across different tasks. LM represents large language model, SM represents small model, such as U-Net, ViT. Text means using language input or not. M-task and Cls denote multi-task and classification, respectively.}
\label{tab:method_comparison}
\begin{tabular}{lccccccc}
\toprule
Method & Venue  & LM/SM & Text & M-task  & Cls & VQA  & Scientific Regression\\

\midrule
Open-Canopy ~\cite{fogel2024open} & CVPR$\prime$25 & SM & \checkmark & \texttimes & \texttimes & \texttimes   & Canopy height \\
\hline
Contextformer~\cite{benson2024multi} &CVPR$\prime$24 & SM & \texttimes & \texttimes & \texttimes & \texttimes   & Geospatial vegetation \\

\hline
\multirow{3}{*}{TorchSpatial ~\cite{wu2024torchspatial} } &\multirow{3}{*}{NeurIPS$\prime$24 } & \multirow{3}{*}{SM} &  \multirow{3}{*}{\texttimes}  & \multirow{3}{*}{\checkmark} &  \multirow{3}{*}{\checkmark} & \multirow{3}{*}{\texttimes}  & Landscape variables, \\ 
 & & & & & & &Urbanization metrics, \\
&  & & & & &  & SDGs indicators\\
\hline

SRMS~\cite{hamilton2024combining} &NeurIPS$\prime$24 & LM & \checkmark & \texttimes & \texttimes & \texttimes   & Species range \\
\hline
LHRS-bot~\cite{muhtar2024lhrs} &ECCV$\prime$24 & LM  & \checkmark & \checkmark & \checkmark & \checkmark & \texttimes \\
\hline
GeoChat~\cite{kuckreja2024geochat} &CVPR$\prime$24 & LM  & \checkmark & \checkmark  & \checkmark & \checkmark  & \texttimes \\
\bottomrule
\end{tabular}
\end{table*}

Against this backdrop, the emergence of Vision Language Models (VLMs)~\cite{ouyang2022training,radford2021learning,liu2024visual,achiam2023gpt} offers promising directions for advancing EO analysis. As illustrated in Figure~\ref{fig:motivation_fig}(a), existing VLM applications in EO focus mainly on perception tasks (e.g., detection) and image content description tasks (e.g., VQA). However, the potential of VLMs for scientific regression tasks, such as predicting environmental attributes, has received limited attention, despite its essential role in EO applications.

To bridge the gap, this work explores the potential of VLMs to jointly tackle both visual description and scientific regression tasks in EO. Compared with traditional methods based on visual contents, VLMs offer unique advantages for solving regression problem,  as illustrated in Figure~\ref{fig:motivation_fig}(b), By integrating multimodal EO data and embedding domain knowledge in language, the VLMs overcome information bottlenecks and conducts regression using more comprehensive information. Despite the potential of jointly modeling regression and generation tasks with VLMs, several key technical challenges are required to be considered: 

\begin{itemize} 
\item \textbf{Lack of instruction-tuning data and evaluation benchmarks:} Existing EO VLM dataset typically emphasizes visual content description, resulting in insufficient fine-tuning data and standardized benchmarks for scientific regression tasks. This limits the models' ability to learn accurate numeric mappings between EO inputs and scientific variables.

\item \textbf{Divergent latent feature requirements:} Generation tasks always focus on describing visual content using features that align with human perception. Scientific regression tasks predict physical properties from raw EO data. They depend on subtle spectral or spatial patterns that may not be visible to the human eye. This difference in feature needs makes it challenging for a single VLM to handle both tasks effectively.

\item \textbf{Error accumulation in multi-step number generation:}  Numerical values are split into multiple discrete tokens during tokenization. Under the autoregressive next-token prediction mechanism, an error in predicting any one token can cascade through subsequent tokens, degrading the accuracy of the final numeric output.

\textbf{}

\item \textbf{Conflicting optimization objectives:} VLMs are trained to predict the next token based on semantic coherence rather than minimizing numeric loss. This mismatch means that optimizing for fluent language generation may undermine the precision needed for accurate numeric regression, making joint fine-tuning challenging.

\end{itemize}

To address these challenges, this paper introduces REO-Instruct, a novel benchmark dataset in the EO domain that uniquely enables the exploration of VLMs’ generation and scientific regression capabilities. 
Specifically, we selected Above Ground Biomass (AGB) estimation and ecological patch counting as representative scientific regression tasks, while VQA and classification as the representative generation task. Around these tasks, the dataset provides extensive text annotations and domain-specific knowledge, including descriptions of land cover types, human activities, and ecological contexts. As illustrated in Figure~\ref{fig:motivation_fig}(c), REO-Instruct highlights the intrinsic relationships between regression targets (e.g., AGB values) and generation objectives (e.g., land cover classes).

In summary, our contributions include:

\begin{itemize}
\item \textbf{Conceptual framework for scientific VLMs in EO:} We introduce a new perspective that integrates content understanding and quantitative regression within Vision–Language Models, guided by a logical-chain design that links semantically and scientifically related tasks. This framework enables coherent multimodal learning and bridges descriptive and numeric reasoning in Earth Observation.

\item \textbf{Creation of REO-Instruct:} We construct \textbf{REO-Instruct}, a large-scale multimodal benchmark that unifies generation and scientific regression tasks in EO. The dataset integrates co-registered Sentinel-2 and ALOS-2 imagery with structured text annotations, forming a cognitively interpretable chain from human activity and land cover to ecological patch counts and above-ground biomass (AGB).

\item \textbf{Comprehensive evaluation and insights:} We benchmark representative VLMs on REO-Instruct with standardized protocols, revealing their current limitations in scientific numeric reasoning and establishing transparent baselines for future development of scientifically grounded VLMs in remote sensing.
\end{itemize}

\section{Related Work}

\begin{figure*}
    \centering
    \includegraphics[width=1.0\linewidth]{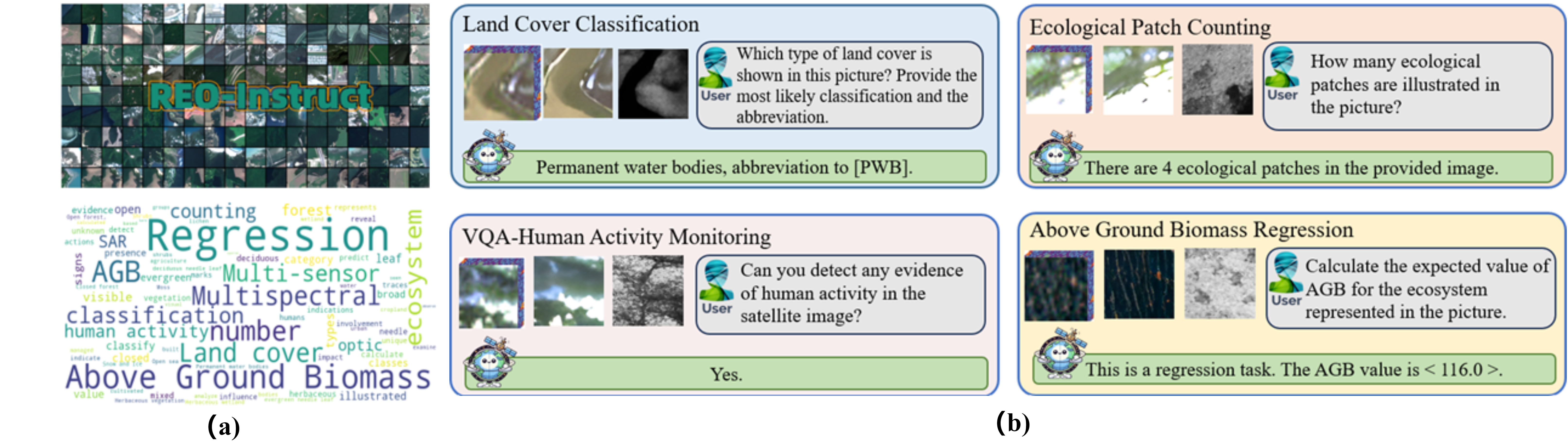}
    \caption{(a). RGB modality examples and word cloud of REO-Instruct benchmark; (b). Screenshots of some image-texts annotation pairs in REO-Instruct benchmark. }
    \label{fig:anno}
\end{figure*}

\label{sec:Related}

\subsection{Vision Language Datasets for EO}
Training VLMs for EO requires specialized image-text datasets, which are typically developed in two main ways: creating datasets from scratch and expanding existing EO datasets.

Creating datasets from scratch involves sourcing raw EO data and pairing it with annotations. For example, RSGPT~\cite{hu2023rsgpt} created 2,500 high-quality RSI text pairs with expert manual captioning. GRAFT~\cite{mall2023remote} linked ground-level image captions from social media with RSIs using geographical tags, obtaining a large dataset without manual captioning. SkyScript~\cite{wang2024skyscript} and LHRS-Align~\cite{muhtar2024lhrs} utilized OpenStreetMap to generate captions, with LHRS-Align further leveraging the language-only LLM Vicuna-v1.5 for caption production based on geographic features.

Expanding existing EO datasets involves transforming existing annotations into textual descriptions. For instance, RemoteCLIP~\cite{liu2024remoteclip} converted object detection annotations into image captions using textual templates, significantly increasing the available training data. Similarly, EarthGPT~\cite{zhang2024earthgpt} adopted a similar template-based approach. The RS5M~\cite{zhang2024rs5m} dataset, currently the largest with 5 million RSIs, employed BLIP2 to generate captions, selecting the best variants using CLIP. GeoChat~\cite{kuckreja2024geochat} constructed question-answer pairs by describing target characteristics, which were subsequently processed by a  LLM. Additionally, SkyEyeGPT~\cite{zhan2024skyeyegpt} combined object detection and VQA datasets to create a multitask dialogue instruction dataset.

These datasets set strong benchmarks for tasks like image captioning, VQA, and visual grounding, focusing mainly on interpreting image content. However, despite laying a solid foundation for future research, they still underutilize the potential of multimodal EO data collected from diverse sensors, particularly in addressing scientific regression challenges.

\subsection{Scientific Regression Tasks in EO}

Scientific regression tasks play a key role in EO, enabling essential scientific applications across climatology, ecology, and geophysics. Recent advances have driven significant progress in EO-related regression tasks~\cite{benson2024multi,hamilton2024combining,wu2024torchspatial}. Common tasks include species range estimation, geospatial vegetation forecasting, population density regression, forest cover prediction, nightlights intensity estimation, elevation mapping, and notably, above ground biomass (AGB) estimation~\cite{lang2023high,sialelli2024agbd}. Several advanced methods have emerged addressing these tasks with multimodal data, as summarized in Table~\ref{tab:method_comparison}. 
Classic regression models in EO typically rely on satellite imagery~\cite{li2020forest,laurin2018above,rodda2024lidar}, complemented by ground-truth data for model training and validation. However, relying solely on EO imagery for scientific regression faces inherent information bottlenecks, as some key environmental and anthropogenic factors may be underrepresented. Incorporating domain-specific knowledge or intermediate interpretation results as textual inputs or embeddings could unlock new possibilities for deeper understanding and reliable prediction.

\section{REO-Instruct Benchmark}

Developing a unified EO-VLM for both generation and scientific regression tasks is crucial yet currently hindered by  lacking benchmarks. To bridge this gap, we propose REO-Instruct, a large-scale multimodal benchmark integrating EO imagery with domain-rich text annotations, supporting comprehensive model assessment.

\subsection{Logical Chain Construction and Task Selection}
\label{sec:logical chain}
Given the diverse range of EO-based generation and scientific regression tasks, constructing a logical chain within the benchmark requires careful consideration guided by the following principles: 1) \textbf{The output of the generation and regression tasks should be derivable from the input EO data},  improving the reliability and interpretability of the results; 2) \textbf{Generation and regression tasks must share tightly-coupled logical relationships}. It is unreasonable and unnecessary to use a unified framework to handle unrelated tasks.
3) \textbf{Text annotations must be clear, concise, and professionally structured}. Recent studies~\cite{tang2024understanding,lewkowycz2022solving,song2024omnipred,vacareanu2024words} demonstrate that while perception tasks are relatively insensitive to minor semantic variations, slight textual perturbations can dramatically affect regression predictions in LLMs. Therefore, clear annotations are critical to mitigating unintended variability.

Based on these guidelines, we specifically focus on forest environmental monitoring, establishing a concise logical chain that incorporates critical and necessary factors, including human activities (anthropogenic impacts), ecological patch counts (reflecting biodiversity), land cover classification and biomass measurements.

\subsection{EO Data Collection Principles and Overview }

To ensure the dataset is comprehensive and representative, we follow these three main principles: 1) EO image modalities must be sufficient and necessary for precise forecasting into surface and vegetation characteristics;  2) The element of EO image data  must be balanced, diverse, and representative. It should cover various land cover types, AGB values, human influences, and geographic distributions to build generalizable models. 

REO-Instruct benchmark leverages the AGBD dataset~\cite{sialelli2024agbd}, which encompasses imagery collected during the years 2019-2020. Inspired by prior work~\cite{li2020forest,laurin2018above,rodda2024lidar},
the proposed benchmark includes three types of EO data: multispectral (MS) images, RGB three-channel optical images, and Synthetic Aperture Radar (SAR) images. The multispectral data comes from Sentinel-2 L2A, covering 13 spectral bands at a 10-meter spatial resolution. We extracted bands [4,3,2] from each multispectral image to create corresponding RGB image. The 25-meter resolution SAR data originates from ALOS-2 PALSAR-2 products, featuring both HH and HV polarization back scatter. More detailed information can refer to~\cite{shimada2014new}.  Images from different modalities have been spatially aligned, resulting in 25$\times$25-pixel patches corresponding to a 250m$\times$250m observation area.

\begin{figure}
    \centering
    \includegraphics[width=0.8\linewidth]{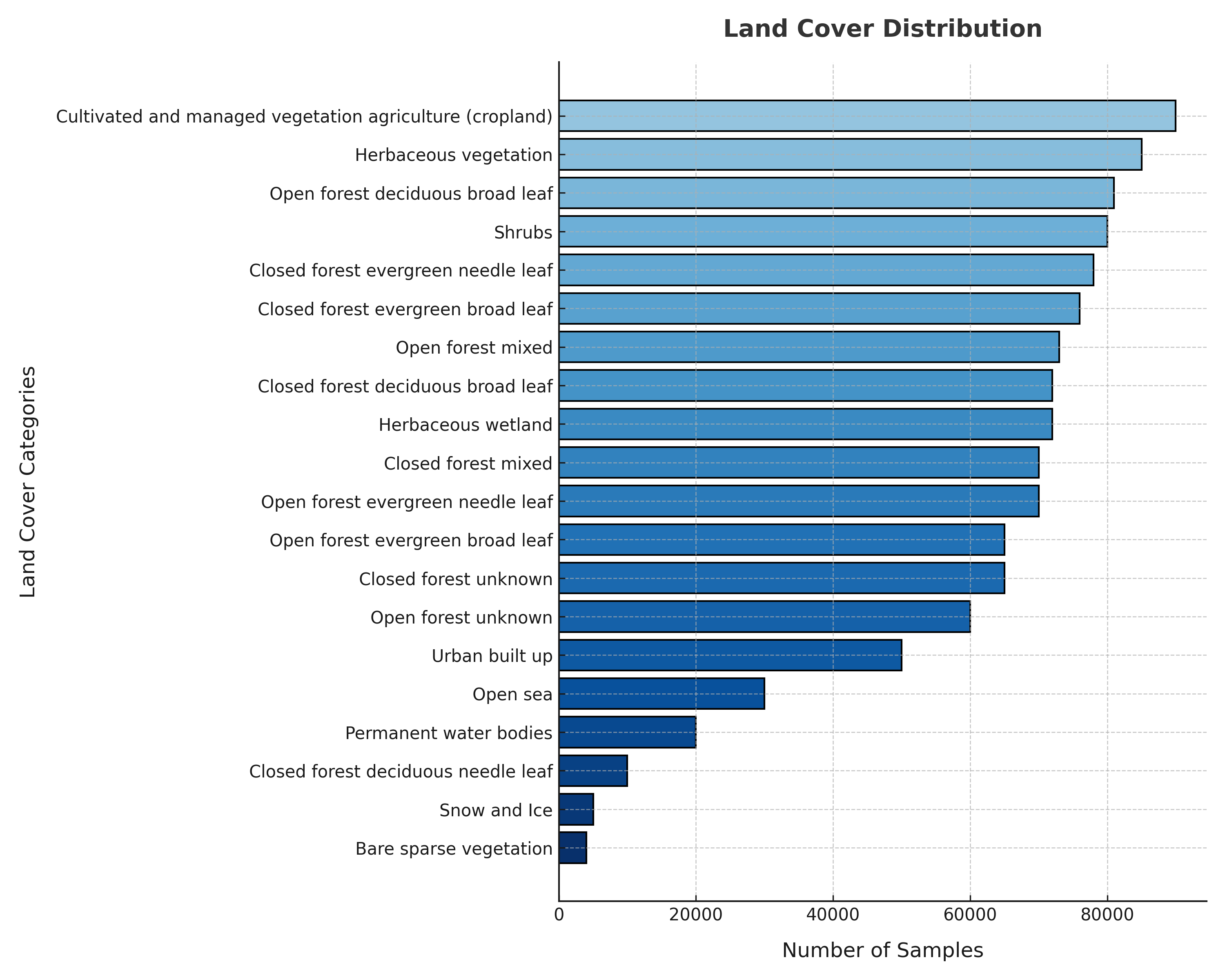}
    \caption{Land cover distribution based on the number of samples in each category. }
    \label{fig:Lc-distribution}
\end{figure}

These EO images captured by different sensors are further enriched with domain-specific text annotations, leveraging land cover data and the GEDI AGB data to establish text annotation. REO-Instruct benchmark comprises a significant volume of image-text pairs, with 1.6 million pairs in training set, approximately 20k pairs in validation set and 36K pairs in testing set. Figure~\ref{fig:anno} shows image examples and focus areas of the proposed REO-Instruct benchmark.

\subsection{Text Annotations in REO-Instruct}
Text annotations must incorporate scientific and domain-specific knowledge relevant to generation and scientific regression tasks. To generate the textual component of our dataset, we utilize ChatGPT-4o\footnote{This work invoked ChatGPT-4o, also known as GPT-4o, using OpenAI's official API.}, guided by a carefully designed prompt that ensures the inclusion of pertinent domain knowledge. Following the construction principle, the text annotations in the REO-Instruct benchmark cover several key aspects:

\begin{figure}[H]
    \centering
    \includegraphics[width=0.8\linewidth]{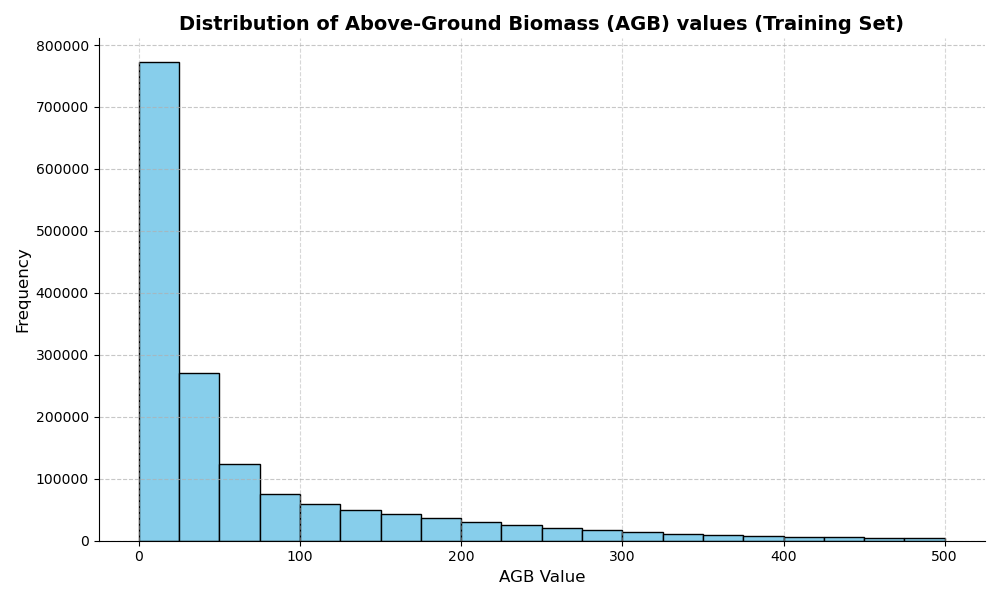}
    \includegraphics[width=0.8\linewidth]{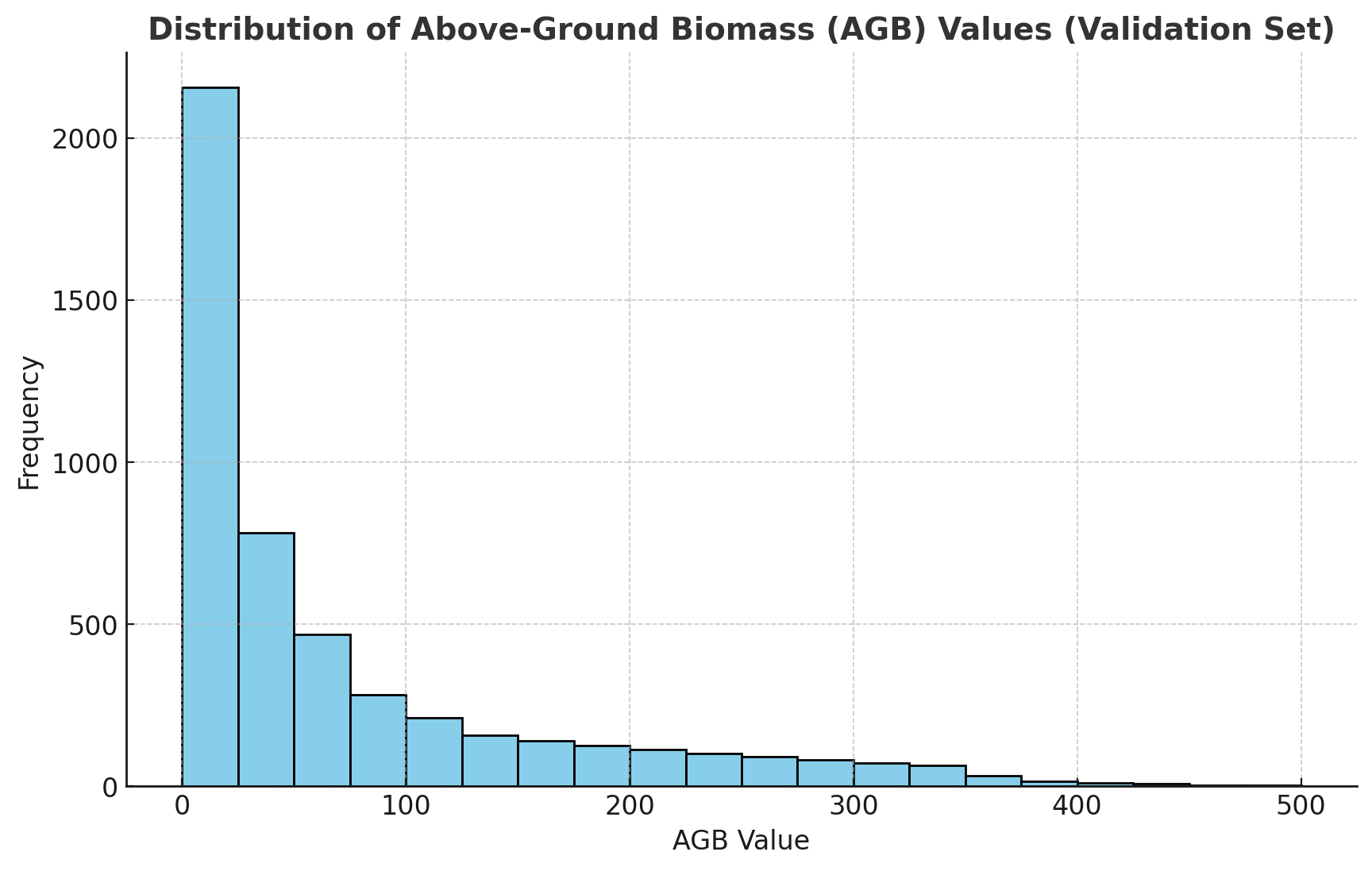}
    \includegraphics[width=0.8\linewidth]{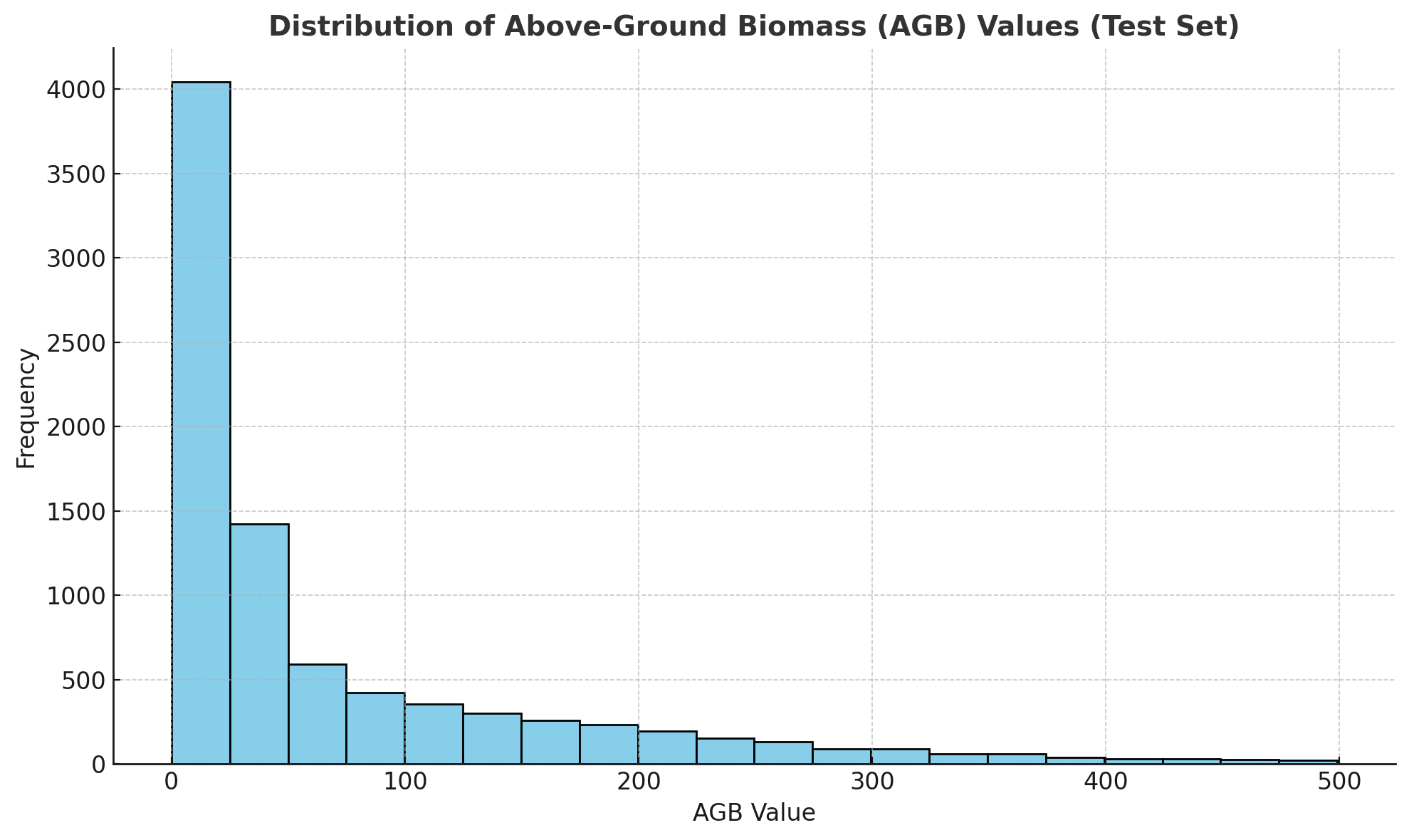}
    \caption{Distribution of Above-Ground Biomass (AGB) values in REO-Instruct. The histogram shows the frequency of AGB values.}
    \label{fig:agb-distribution}
\end{figure}

\begin{itemize}
    \item \textbf{Land Cover Classification:} Text annotations provide detailed descriptions of land cover types based on Copernicus Global Land Cover Layers. Each image is assigned to one of the professional land cover classes, such as \textit{Closed forest, evergreen needleleaf forest}. These labels enable models to learn vegetation structures, land cover compositions, and spatial patterns critical for ecological and environmental analysis. A complete list of more than 20 land cover categories and their distributions in REO-Instruct can be found in Figure~\ref{fig:Lc-distribution}.

     \item \textbf{Ecological Patch Counting:} The annotations provide estimates of the number of ecological patches within each observed area. An ecological patch is defined as a continuous land cover unit with distinct ecological characteristics, such as vegetation type, land use, or habitat features. This annotation reflects vegetation richness and fragmentation, where a higher number of patches indicates greater biodiversity and land cover complexity, offering critical ecological insights.
     
     \item \textbf{VQA-Human Activity Monitoring:} 
     This part of annotations include questions and answers about human-made features like urban structures, agricultural fields. These annotations discuss the potential impact of human activities on natural landscapes, facilitating the study of human-environment interactions such as urban expansion and deforestation-driven land-use change.
     \item \textbf{Above Ground Biomass Regression:} Text annotations further provide ground-truth quantitative estimates of Above Ground Biomass (AGB), reported in megagrams per hectare (Mg/ha). These values allow models to connect textual descriptions with visual inputs, enabling direct supervision for biomass regression tasks and supporting advanced multimodal learning frameworks. Details about AGB value distribution in REO-Instruct are shown in Figure~\ref{fig:agb-distribution}.

\end{itemize}

\begin{table}[t]
  \centering
  \caption{Results (\%) on land-cover classification.}
  \label{tab:classification}
  \small
  \setlength{\tabcolsep}{3pt} 
  \begin{tabular}{l|ccccc}
  \toprule
  Method & Modality & {OA$\uparrow$} & {MA-Pre$\uparrow$} & {MA-Recl$\uparrow$} & {MA-F1$\uparrow$} \\
  \midrule
  Qwen2-VL*  & RGB & 3.77 & 0.38 & 1.14 & 0.57 \\
  ChatGPT-4o & RGB & 3.97 & 12.27 & 3.63 & 5.60 \\
  \bottomrule
  \end{tabular}
  {\footnotesize *: metrics exclude unanswerable queries. Qwen2-VL deems 96.06\% unanswerable.}
\end{table}

\begin{table}[t]
  \centering
  \caption{Results on ecological patch counting task.}
  \label{tab:ecological patch}
  \small
  \setlength{\tabcolsep}{2.5pt} 
  \begin{tabular}{l|ccccc}
    \toprule
    Method & Modality & RMSE$\downarrow$ & MAE$\downarrow$ & R-squared$\uparrow$ & OA (\%)$\uparrow$ \\
    \midrule
    Qwen2-VL* & RGB & 12.74 & 4.30 & -121.79 & 9.75 \\
    ChatGPT-4o & RGB & 5.13 & 4.79 & -18.42 & 2.43 \\
    LLaVA* & RGB & 1.31 & 1.06 & -0.27 & 25.34  \\
    \bottomrule 
  \end{tabular}
  {\footnotesize *: metrics exclude unanswerable queries. LLaVA answers 79.31\% of RGB and 21.20\% of MS queries, while Qwen2-VL only answers 20.31\% questions.}
\end{table}

\subsection{Prompt Design  and Manual Correction}
To guide ChatGPT-4o in generating the most relevant answers while ensuring diversity in the responses, we have carefully designed a large number of prompts. For instance, to ensure the accuracy of land cover category descriptions, we restrict the responses to be selected from a predefined set of available categories. To ensure the diversity of questions and answers, we have set up more than 100 templates. Each conversation is generated by randomly selecting a template, while also introducing new variations to create unique responses. Additionally, we categorize the questions explicitly to ensure that the regression head is activated solely for regression-related tasks. Meanwhile, the generation head does not produce irrelevant content, which could interfere with the regression head's decision-making.

\definecolor{rowcolor1}{RGB}{220, 230, 255}
\definecolor{rowcolor2}{RGB}{200, 215, 250}
\definecolor{rowcolor3}{RGB}{180, 200, 245}
\definecolor{rowcolor4}{RGB}{160, 185, 240}

\begin{table}[t]
  \centering
  \caption{Results on Human Activity Monitoring  task.}
  \label{tab:vqa}
  \small
  \setlength{\tabcolsep}{14pt} 
    \begin{tabular}{l|cc}
    \toprule
    \textbf{Model} & \textbf{Modality} & \textbf{Accuracy (\%)} $\uparrow$ \\
    \midrule
    Owen2-VL & RGB & 21.52 \\
    ChatGPT-4o & RGB & 33.79 \\
    LLaVA & RGB & 44.92 \\
    GeoChat & RGB & 45.03 \\
    LHRS-Bot & RGB & 47.87 \\
    LLaVA & MS & 48.08 \\
    \bottomrule
    \end{tabular}
\end{table}

\begin{table}[h]
  \centering
  \caption{Results on AGB regression task.}
  \label{tab:agb}
  \small
  \setlength{\tabcolsep}{10pt}
  \begin{tabular}{@{}l |l c c c@{}}
    \toprule
    Method        & Modality & RMSE$\downarrow$ & MAE$\downarrow$ & $R^2\uparrow$ \\
    \midrule
    LLaVA$^{\dagger}$     & RGB & 116.60 & 67.90 & -0.51 \\
    U-Net          & MS  &  81.27 & 49.19 &  0.32 \\
    LLaVA*$^{\dagger}$    & MS  & 115.74 & 67.49 & -0.45 \\
    \bottomrule
  \end{tabular}
  {\footnotesize
  \\*: Unanswerable cases excluded; $^{\dagger}$: Models fine-tuned on REO-Instruct. LLaVA*$^{\dagger}$ only counted 89.31\% of questions with definite answers.}
\end{table}

Because automatic generation can introduce errors or drift, we added a manual correction step to keep our annotation accurate and reliable. First, every model-assisted QA pair goes through an automated check. A script compares each annotation against trusted sources, like the \textit{Copernicus Global Land Cover Map}, and fixes any mismatches. For example, if the land cover map marks an area as “urban” but ChatGPT-4o’s label doesn’t mention human activity, we correcte the model-generated labels to match. Any entries that remain unclear are removed. Next, senior experts review the cleaned test and validation sets by hand. They read through each annotation to confirm facts and ensure terminology is consistent with domain standards. This two-step process automatic filtering followed by expert review helps us deliver a benchmark that’s both scientifically sound and easy to trust.

\section{Experiments}

\begin{figure*}
    \centering
    \includegraphics[width=.95\linewidth]{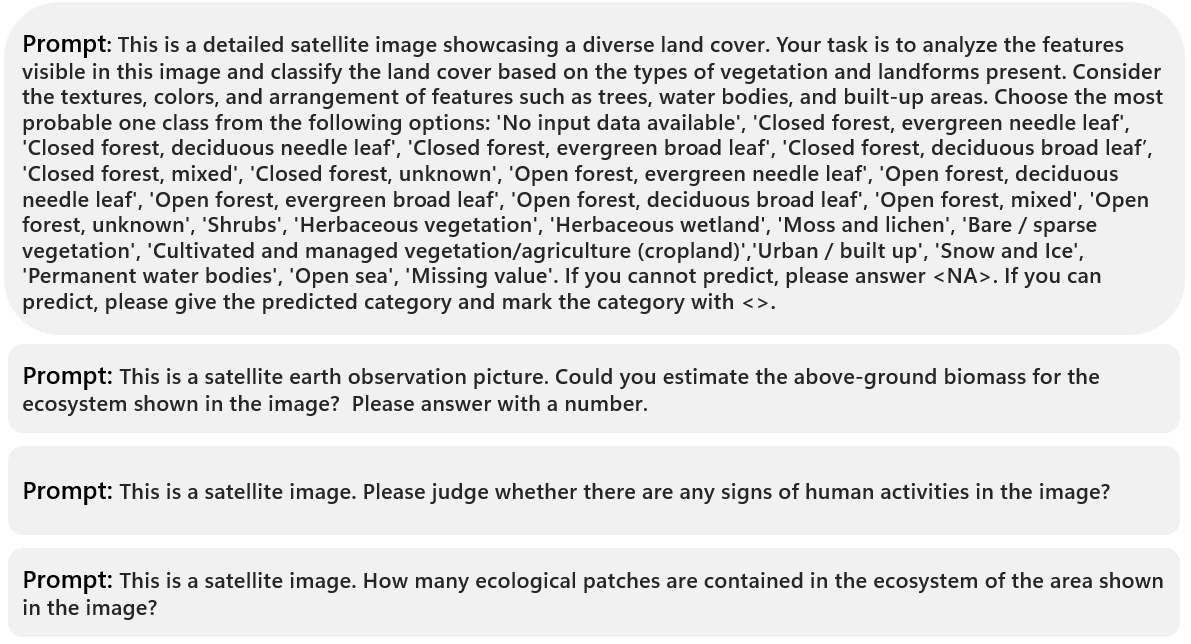}
    \caption{Test guiding prompts for compared methods.}
    \label{fig:prompts}
\end{figure*}

We present results on the REO-Instruct benchmark to evaluate the performance of representative VLMs. Each downstream test subset contains approximately 8.6K unique samples, ensuring no image overlap.

For comparison, we include both domain-specific and general-purpose VLMs. Specifically, we evaluate two EO focused large models, GeoChat~\cite{kuckreja2024geochat} and LHRS-Bot~\cite{muhtar2024lhrs}, alongside general-purpose VLMs including LLaVA-1.5-7B~\cite{liu2024visual}, Qwen2-VL-7B~\cite{wang2024qwen2}, and ChatGPT4o~\cite{achiam2023gpt}. In addition, we consider the results obtained when using different modalities of EO imagery as inputs to REO-VLM. This initial exploration aims to assess how varying input modalities influence inference accuracy. For GeoChat, we modified its \textit{clip\_interpolate\_embeddings} component to adapt to the image resolution of REO-Instruct. 
Unless otherwise specified, for the comparison models, we use the official 7B versions. Following common evaluation practices in EO-VLM research~\cite{li2024vision,kuckreja2024geochat,muhtar2024lhrs}, we directly evaluate these publicly released models without any additional training on our dataset.  To ensure fairness during evaluation, we further provided guiding prompts that explained the questions and offered a range of possible answers to other compared methods.

\subsection{Guiding Prompts}
To define the scope of our questions and expected answers, we provide guiding prompts during testing for all comparison algorithms, except for our proposed method. These prompts shown in Figure \ref{fig:prompts} assist VLMs in better understanding and addressing multiple tasks.

\subsection{Experimental results}

We present experimental results across four representative tasks: land cover classification (Table~\ref{tab:classification}), ecological patch counting (Table~\ref{tab:ecological patch}), VQA-based human activity monitoring (Table~\ref{tab:vqa}), and above-ground biomass (AGB) regression (Table~\ref{tab:agb}) on the REO-Instruct benchmark. Building on these quantitative results, we summarize three key observations.
\begin{itemize}
    \item In the human-activity monitoring task, MS inputs consistently outperform RGB imagery, and EO-specific VLMs (GeoChat, LHRS-Bot) achieve higher accuracy than general-purpose models, underscoring the benefit of spectral cues and domain-specialized pre-training. 
    \item For forest ecology–oriented questions, general-purpose open-source and closed-source VLMs exhibit very low answer accuracy, revealing a substantial gap in forestry and ecological knowledge that cannot be bridged by generic web-scale corpora alone.
    \item In the regression tasks, all compared methods yield negative $R^{2}$ values under both RGB and MS settings, indicating that they fail to learn meaningful numerical relationships from EO inputs and remain far from reliable for EO-driven quantitative estimation.
\end{itemize}

\section{Conclusion}

This work benchmarks VLMs in EO for forest ecological analysis through the introduction of \textbf{REO-Instruct}, a multimodal benchmark designed to align descriptive and quantitative tasks within a unified framework. Experiments across representative tasks demonstrate that while current VLMs excel in content understanding, they still face significant challenges in numerical reasoning and scientific regression. 
These findings highlight the necessity of regression-aware modeling strategies and multimodal data integration for advancing scientific applications of VLMs. 
Overall, this work establishes a standardized foundation for benchmarking and developing next-generation \textbf{scientific VLMs in forest ecological analysis}, with future efforts directed toward extending this framework to broader geoscientific domains.

\section{Acknowledgments}
The work of Xizhe Xue is funded by the Technical University of Munich Global Postdoc Fellowship. This work is also supported by Munich Center for Machine Learning.

\bibliography{aaai2026}

\end{document}